\documentclass[runningheads]{llncs}
\usepackage[T1]{fontenc}
\usepackage{graphicx}
\usepackage{orcidlink}
\usepackage{hyperref}
\usepackage{graphicx}
\usepackage{booktabs}
\usepackage{appendix}
\usepackage{xcolor}
\definecolor{TealBlue}{HTML}{1E5B67}      
\definecolor{DarkerTealBlue}{HTML}{287D8E}  
\definecolor{CyanBlue}{HTML}{2A96A5}        
\definecolor{SkyBlue}{HTML}{4E9ABF}         
\definecolor{SlateBlue}{HTML}{5778A3}       
\definecolor{Indigo}{HTML}{6B5C94}          
\definecolor{DarkPurple}{HTML}{854F86}      
\definecolor{MagentaPurple}{HTML}{A3476D}    
\definecolor{CrimsonRed}{HTML}{D13C4F}       
\definecolor{DeepOrange}{HTML}{F07814}       

\begin{document}

\title{Have Large Vision-Language Models Mastered Art History?}
\titlerunning{Have Large Vision-Language Models Mastered Art History?}
%

\author{Ombretta Strafforello\inst{1}\orcidlink{0000-0002-5258-8534} \and
Derya Soydaner\inst{2}\orcidlink{0000-0002-3212-6711} \and
Michiel Willems\inst{1}\orcidlink{0009-0008-6548-2334} \and
Anne-Sofie Maerten\inst{1}\orcidlink{0000-0003-2710-0936} \and
Stefanie De Winter\inst{1}\orcidlink{0000-0002-7551-2444}}
\authorrunning{O. Strafforello et al.}
%
\institute{KU Leuven, Leuven, Belgium \\
\email {\{ombretta.strafforello, michiel.willems, annesofie.maerten, stefanie.dewinter\}@kuleuven.be}  \and
Leiden University, Leiden, The Netherlands \\ 
\email{d.soydaner@liacs.leidenuniv.nl}} 
\maketitle              
\begin{abstract}
The emergence of large Vision-Language Models (VLMs) has established new baselines in image classification across multiple domains. We examine whether their multimodal reasoning can also address a challenge mastered by human experts. Specifically, we test whether VLMs can classify the style, author and creation date of paintings, a domain traditionally mastered by art historians. Artworks pose a unique challenge compared to natural images due to their inherently complex and diverse structures, characterized by variable compositions and styles. This requires a contextual and stylistic interpretation rather than straightforward object recognition. Art historians have long studied the unique aspects of artworks, with style prediction being a crucial component of their discipline. This paper investigates whether large VLMs, which integrate visual and textual data, can effectively reason about the historical and stylistic attributes of paintings. We present the first study of its kind, conducting an in-depth analysis of three VLMs, namely CLIP, LLaVA, 
and GPT-4o, evaluating their zero-shot classification of art style, author and time period. Using two image benchmarks of artworks, we assess the models' ability to interpret style, evaluate their sensitivity to prompts, and examine failure cases. Additionally, we focus on how these models compare to human art historical expertise by analyzing misclassifications, providing insights into their reasoning and classification patterns.
\keywords{Style classification \and Multimodal learning \and AI in art history.}
\end{abstract}
\section{Introduction}
Classification in art history involves a range of attributes, including authorship, period, and geographical origin, all of which help place an artwork within its historical and cultural context. 
Additionally, factors such as medium, technique, and iconography provide insights into the methods and symbolism used by artists. 
Together, these elements form the basis for analyzing and categorizing art. 
However, the classification of an artwork's attributes, 
which encapsulates numerous visual and historical features, is a complex task. Art style classification, ranging from highly realistic to purely abstract, as illustrated in Appendix~\ref{Appendix A}, is particularly challenging, even for human experts.  
Automating art style classification could yield significant benefits, such as categorizing newly discovered artworks, efficiently managing large collections, verifying existing style labels, and improving the quality of datasets --- all while conserving time and resources.
This has motivated computer vision researchers to develop neural networks for the classification of artworks and, especially, their style.

Recently, large VLMs have revolutionized computer vision by leveraging their multimodal nature to interpret both visual and textual input. Typically trained to match images with text descriptions or answer questions about an image content, VLMs benefit from the semantic richness of language, compared to the 
single category labels used to train traditional image classifiers, 
helps VLMs capture more nuanced image characteristics, 
such as object location within an image \cite{dorkenwald2024pin}, the geolocation \cite{mendes2024granular} or the image aesthetic qualities  
\cite{lin2024vila}.
Their extensive multimodal training on large-scale datasets scraped from the internet, make VLMs successful in various image domains, ranging from remote sensing \cite{al2023vision} and medical imagery \cite{bazi2023vision} to cartoon captioning \cite{cao2024predicting}. The success of VLMs has led us to ask: To what extent can VLMs solve the complex task of art classification? 

%
In this paper, we present the first evaluation of VLMs for this purpose. We utilize two datasets, WikiArt and JenAesthetics, and evaluate three VLMs: CLIP, LLaVA, 
and GPT-4o. While newer models have since been introduced, the selected models provide a representative understanding of current VLM performance in art historical analysis. We conduct a rigorous evaluation of their zero-shot performance in predicting art historical attributes, including art style, author, and time period of paintings, using different prompting methods. While fine-tuning or in-context learning are promising directions, we deliberately focus on evaluating zero-shot inference. This approach \textit{1)} provides a foundational understanding of VLMs' performance before designing improvements, \textit{2)} keeps our evaluation closely aligned with typical VLM usage by non-expert users (e.g., through ChatGPT), and \textit{3)} tests whether VLMs' strong zero-shot performance across multiple image domains holds with artworks.

Beyond measuring classification performance, we investigate 
the impact of adding contextual information in the
text prompts and analyze failure cases to assess the extent of their multimodal reasoning about artworks. 
Our findings show that while CLIP, LLaVA, and particularly GPT-4o exhibit a certain level of accuracy, some errors do not align with the acceptable standards of art history experts. We conclude that current VLMs are not yet suitable for use in this field without expert oversight. Our code will be made public upon acceptance.

\section{Related Work}
\subsection{Automating the Classification of Artworks}

Research on automatic art style classification has established important baselines. Early work explored photographic style recognition \cite{karayev2013recognizing} 
and optimized similarity measures between paintings \cite{saleh2015large}. Later studies leveraged Convolutional Neural Networks (CNNs) for fine-art painting classification \cite{bar2015classification,tan2016ceci,lecoutre2017recognizing,elgammal2018shape,cetinic2018fine,sandoval2019two,li2025enhanced}. Other methods, such as a boosted ensemble of Support Vector Machines, recognize artistic movements of paintings \cite{florea2018artistic}. Various CNNs have achieved accuracies of 59.01\% \cite{zhong2020fine} and 68.55\% \cite{Menis2020Deep} on WikiArt\footnote{https://www.wikiart.org/}. Zhao \textit{et al.} \cite{zhao2022big} applied transfer learning to classify all 27 styles in WikiArt, achieving the current state-of-the-art with an accuracy of 71.24\%. Appendix~\ref{Appendix B} summarizes prior results on WikiArt.

Closely related to automating artwork classification, Large Language Models (LLMs) have recently been used including formal art analysis via proxy learning from visual concepts such as color \cite{kim2022formal}, automating art analysis \cite{khadangi2025cognartive}, and zero-shot classification using descriptive prompts \cite{tojima2025zero}. However, these approaches rely solely on text. In contrast, we explore VLMs that process both image and text data, enabling direct visual analysis. We evaluate their ability to predict an artwork's style, author, and time period, providing a comprehensive approach.

\subsection{Vision-Language Models}
VLMs have found wide applicability in tasks like image captioning, visual question answering (VQA), and zero-shot learning (see \cite{bordes2024introduction} for an overview). VLMs
differ from traditional deep learning models by enabling multimodal learning, where visual and textual data are processed together, leading to a deeper contextual understanding. In art research, CLIP \cite{radford2021learning} has been used to classify metadata in art-historical images \cite{conde2021clip}. Its features have also been applied for content and style disentanglement \cite{wu2023not}. Other studies have used VLMs for qualitative results on style classification \cite{lin2024vila} and iconographic concept recognition \cite{springstein2024visual}. 

A related work is GalleryGPT \cite{bin2024gallerygpt}, which generates formal analyses of paintings using short descriptive paragraphs by fine-tuning LLaVA. In contrast, we explore VLMs for style prediction, conducting a comprehensive evaluation across multiple models. We assess their performance in predicting artistic styles, time periods, and authors. Another work, ArtGPT-4 \cite{yuan2023artgpt} interprets and articulates emotions evoked by paintings. To our knowledge, we present the first performance analysis of VLMs in style, author, and time period classification, establishing a foundation for future research in multimodal reasoning and art analysis.

\section{Predicting Artistic Attributes with VLMs}
\subsection{Chosen VLMs} 

\textbf{CLIP} \cite{radford2021learning} is a VLM trained to match correct pairings of images and natural language descriptions using separate encoders for image and text. In our tests, both encoders are Transformer-based models \cite{dosovitskiy2020image,vaswani2017attention}. CLIP employs a contrastive objective to bring matching image-text pairs closer in embedding space and push non-matching ones apart. Trained from scratch on 400 million image-text pairs collected from the internet, CLIP generalizes well across visual concepts and language queries. We utilize the official OpenAI implementation\footnote{\url{https://github.com/openai/CLIP}}.

\noindent \textbf{LLaVA} (Large Language and Vision Assistant) \cite{liu2024visual} is a large multimodal model for image understanding and conversation, capable of generating text and answering image-related questions. It is trained in two stages: (1) feature alignment, using 558K subset of the LAION-CC-SBU\footnote{https://huggingface.co/datasets/liuhaotian/LLaVA-Pretrain} dataset to connect a frozen pretrained vision encoder with a frozen LLM; and (2) visual instruction tuning, using 150K GPT-generated instruction data and approximately 515K VQA samples. This training enables LLaVA to follow multimodal instructions. This multimodal chatbot can perform well in tasks that require a deep visual and language understanding, such as image captioning and VQA. We use llava-hf/llava-1.5-7b-hf checkpoints hosted by Hugging Face. 

\noindent \textbf{GPT-4o} \cite{achiam2023gpt} is an advanced, closed-source multimodal model with a Transformer-based architecture, pretrained to predict the next token in a sequence. Trained on a massive amount of image-text pairs from public sources and licensed data providers, it achieves human-level performance on various benchmarks.
In our tests, we use the OpenAI API\footnote{ \url{https://platform.openai.com}}. Due to its usage cost, we applied GPT-4o more selectively compared to the freely accessible CLIP and LLaVA.

\subsection{Datasets}

\textbf{WikiArt} is the largest public collection of digitized artworks, with 81,449 paintings by 1,119 artists, ranging from the 15th century to contemporary times. These paintings represent 27 styles (e.g., abstract, impressionism) and 45 genres (e.g., landscape, portrait), with some paintings annotated with their creation year. It is divided into a training set (57,025 images) and a validation set (24,421 images). We evaluate model performance on the validation set. Since not all the paintings have their creation year annotated, we use the subset of 7,759 paintings dated after 1900 and we refer to this set as WikiArt \textit{Validation Subset} (VS).

\noindent \textbf{JenAesthetics} \cite{amirshahi2013,amirshahi2013a,amirshahi2014} consists of 1,628 high-resolution scans of oil paintings from museums, covering 11 art periods and 410 artists. It encompasses a comprehensive variety of artistic themes. Due to some broken URLs in the original dataset, we obtained 1,576 out of the 1,628 images.

\subsection{Zero-Shot Prediction} 
\begin{figure}[ht!]
    \centering
    \includegraphics[width=.95\textwidth]{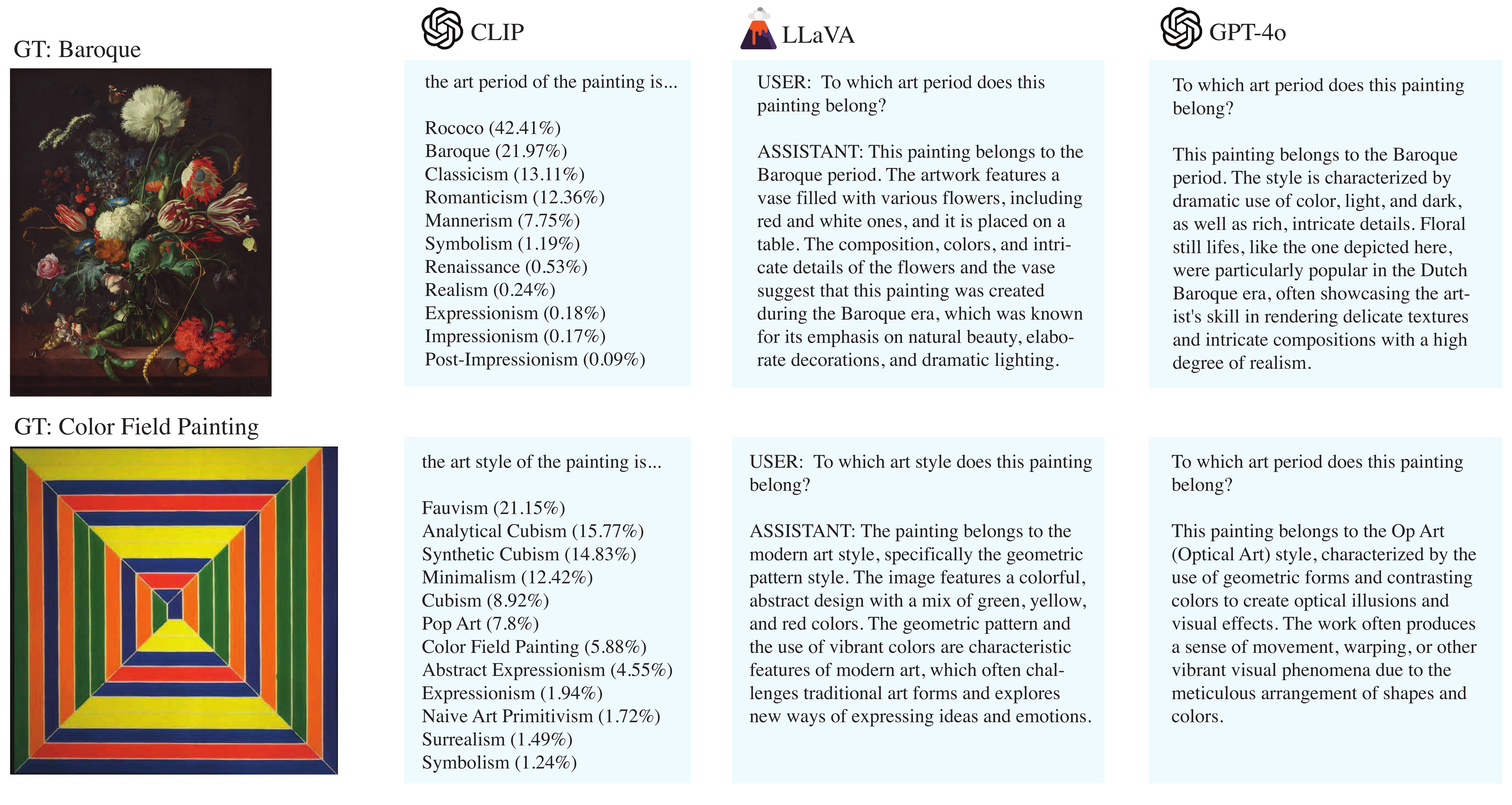}
    \caption{Example of style classification with VLMs on the paintings \textit{Vase of Flowers}, by Jan Davidsz de Heem, c. 1660 and 
    \textit{Color Maze} (1966) by Frank Stella.
    While LLaVA and GPT-4o correctly classify the style of \textit{Vase of Flowers} and CLIP commits a reasonable mistake, all models fail to recognize \textit{Color Maze} as a Color Field painting. 
    }
    \label{fig:Figure1}
\end{figure}

We prompt the VLMs in various ways to predict three attributes of a painting: 
art style, author and time period. These predictions are \textit{zero-shot}, meaning that the models have not been trained for these specific tasks. Figure~\ref{fig:Figure1} illustrates how we prompt VLMs for art style classification and provides examples of the model outputs. 
We assess whether the art style predictions can be improved by enriching the prompts with contextual information, incorporating the year and author into the prompts. We perform these tests using LLaVA.
We also ask the model to predict the time period and the author of a painting. For the time period, we simplify the task by evaluating whether the output falls within the correct \textit{decade} or \textit{century}, instead of the exact year. 

\subsection{Evaluation}

To evaluate CLIP's predictions, we identify the art style, author, and year whose text embeddings are closest to the image embedding and compare them to the ground-truth labels. For language assistants like LLaVA and GPT-4o, which generate varying text lengths, we analyze the output texts in a post-processing step. For this analysis, we deploy the pre-trained Llama 3.2 language model (3B parameters) \cite{meta2024llama} to extract predicted style, author, and time period from the textual description output by the VLMs. Finally, we map the predictions to the exact ground-truth labels using the Jaro-Winkler text distance metric \cite{jaro1989advances,winkler1990string}. 
This step is necessary when the output is correct but not identical to the exact wording in the ground-truth, e.g., “\textit{Neoclassical}” vs. “\textit{Neoclassicism}”. Predictions without a ground-truth match are left unchanged for detailed examination.

\section{Results}
\subsection{Zero-Shot Art Style Prediction}

\subsubsection{Performance evaluation.} 
An overview of the style classification accuracies is provided in Table~\ref{tab:style_prediction_new}. Remarkably, all three VLMs performed significantly better than random chance, even without being specifically trained or fine-tuned to predict art style. 
The highest accuracy is achieved by GPT-4o, which scores 53.42\% on WikiArt and 54.35\% on JenAesthetics. 
The success of GPT-4o can be explained by its massive training process. Nevertheless, a gap remains between the VLMs and the state-of-the-art style classification method, \textit{Big Transfer learning} \cite{zhao2022big}, which correctly classifies 71.24\% of WikiArt's paintings across all 27 classes. This indicates that further improvements are needed before VLMs can be safely deployed for art style classification.  

\begin{table}[!t]
\caption{Zero-shot style prediction accuracy by CLIP, LLaVA, and GPT-4o on WikiArt Validation Subset (VS) and JenAesthetics. The wording \textit{art period} is used in the prompt for the JenAesthetics dataset, to better align with its annotations. For the other datasets, \textit{art style} is used.}
\centering
\fontsize{10}{10}\selectfont
\resizebox{\textwidth}{!}{%
\begin{tabular}{l|l|c|c}
\toprule
\textbf{Model} & \textbf{Prompt} & \begin{tabular}[c]{@{}c@{}}WikiArt Valid.\\ Accuracy (\%)\end{tabular} & \begin{tabular}[c]{@{}c@{}}JenAesthetics\\ Accuracy (\%)\end{tabular} \\
\midrule
CLIP & \textit{the art (style)/(period) of the painting is [art\_style]} & 24.57 & 36.04 \\
LLaVA & \textit{To which art (style)/(period) does this painting belong?} & 28.32 & 37.02 \\
GPT-4o & \textit{To which art (style)/(period) does this painting belong?} & 53.42 & 54.35 \\
\bottomrule
\end{tabular}%
}
\label{tab:style_prediction_new}
\end{table}

To better understand the performance of the VLMs, we examine the F1 score for each style, as illustrated in Appendix~\ref{Appendix F}. The styles with the highest F1 scores across the datasets and models include Baroque, Impressionism, Pointilism, and Pop Art, followed closely by Abstract Expressionism and Romanticism. In contrast, Action Painting, Analytical and Synthetic Cubism, and New Realism tend to have lower scores. The variation in F1 scores reflects the duration of these style movements and the number of paintings produced within them.

Impressionism is among the best and most frequently predicted styles, especially by LLaVA (15.93\% of WikiArt's paintings). This might be due to the popularity of Impressionism, which gained early global acclaim and appeals to all social classes. Its high market value, ongoing educational efforts, and enhanced accessibility through modern technology also contribute to its enduring appeal \cite{snider2001lasting,hook2012ultimate,dombrowski2023impressionism}. In predicting Impressionism, GPT-4o achieves the highest F1 score of 0.71 on WikiArt, 
followed by CLIP (0.53) and LLaVA (0.5).  

The vocabulary of LLaVA is mostly wide and accurate, but it lacks awareness of certain art styles; for instance, it never predicts Mannerism. Additionally, when predicting the art period, LLaVA outputs incorrect labels like Egyptian, Medieval, Modern alongside more conventional art periods such as Baroque, Renaissance and Impressionism.

An overview of the styles predicted by GPT-4o in JenAesthetics, compared to the ground-truth, is available in Appendix~\ref{Appendix C}. Notably, GPT-4o provides a finer-grained classification than the dataset labels; for example, it distinguishes Italian, Venetian, Spanish, and English Renaissance from the overarching `Renaissance' category. The model achieves an overall accuracy of 54.35\%. However, when the predicted styles are grouped according to their overarching stylistic period, the accuracy rises to 62.79\%, with `Modern Art and 20th Century Movements' and  `Post-Impressionism' achieving the highest (75\%) and lowest (40.1\%) accuracies, respectively. 
%
%
%
%
Despite being trained on massive amounts of data, GPT-4o does make mistakes.
Examining these misclassifications provides insights into GPT-4o's multimodal reasoning and its challenges in distinguishing closely related artistic styles, which we elaborate on in the following sections.

\subsubsection{Misclassifications.} 

Despite the relatively low accuracies (24.57\% on WikiArt, 36.04\% on JenAesthetics), CLIP makes sensible mistakes. It confuses Baroque, Rococo, Classicism, and Romanticism; Impressionism, Post-Impressionism, and Expressionism; Color Field Painting, Abstract Expressionism and Minimalism -- styles that are temporally close and share significant visual similarities.
CLIP predicts Cubism mostly as Analytic Cubism. Expressionism, Surrealism, Symbolism and Synthetic Cubism are frequently misclassified as Analytic Cubism. These styles all emerge within the same historical period and exhibit certain visual similarities. Misclassifying Expressionism as Fauvism is understandable due to their shared stylistic traits. A similar pattern occurs with Renaissance styles, which are also conflated. 
Realist paintings misclassified as Romanticist can be explained by a number of factors, including training set bias and similar subject matter. 
The most unexpected error by CLIP is predicting Contemporary Realism instead of the correct styles Impressionism, Romanticism, Realism and Baroque. This mistake might be due to `contemporary' and `realism' are misunderstood as adjectives related to the visual appearance of the subject matters, instead of art style classes. We include the confusion matrices in the Appendices.

We observe that LLaVA confuses Impressionism with several artistic movements in WikiArt, including Post-Impressionism, Realism, Expressionism, Symbolism and Art Nouveau Modern. 
This might be due to Impressionism being prevalent in LLaVA's training data. Similarly, we find that LLaVA is biased towards predicting Romanticism more frequently than other art styles, with 41\% of Romanticism predictions in JenAesthetics. 
Additionally, LLaVA classifies Minimalism paintings as Expressionism, Romanticism, and Realism, which is an anomalous result given the minimal overlap between these styles. Overall, our findings show that LLaVA yields more inconclusive results than CLIP.

GPT-4o makes reasonable predictions across all genres, with the greatest accuracy in Ukiyo-E, Baroque and Impressionism. However, it frequently mislabels works as Baroque--for instance, many Rococo paintings in JenAesthetics. One notable error is misclassifying Frank Stella’s \textit{Color Maze} as Op Art (Optical Art), shown in Figure \ref{fig:Figure1}. Although Stella’s work shares some characteristics with Op Art, such as geometric patterns and contrasting colors, the artist’s intention was not to create the visual vibrations typical of the Op Art style. A closer examination of this piece, which is an example of Color Field Painting, reveals intentional irregularities meant to distinguish it from Op Art.
Similarly, all paintings by Vuillard--typically Post-Impressionist-- are labeled Baroque, likely due to shared texture and fabric details.  
In some cases, GPT-4o fails to predict any style label or outputs `Fresco', which is a technique rather than a style.

\subsubsection{Reasoning Abilities of VLMs vs. Art Historians.}
We further analyze the VLMs' reasoning abilities 
under the oversight of art history experts. Notably, the language assistants attempt to justify the style label, without being prompted to do so. However, their reasoning is often flawed. They often describe the scene or elements that are irrelevant to the style (e.g., \textit{`the scene also includes a man wearing a top hat, which is a characteristic element of Romanticism paintings'}). In other cases, the models cite stylistic elements that could fit into many styles. For example, they claim `the depiction of everyday scenes' is typical for works in the Dutch Golden Age, Northern Renaissance, Impressionism, Realism, Baroque, Rococo and Romanticism. Given that this argument fits many styles, it is not a strong cue for style classification. The reasoning of both language assistants seems inherently different from those of art historians. 
This is evident in the misclassifications of Manet's paintings: Manet created several works that are considered `Realism' before becoming a key figure in the rise of the style `Impressionism'. As a result,  GPT-4o labels \textit{all} of his works as Impressionism. Art historians differentiate Manet's works by observing stylistic elements, whereas GPT-4o seems to lack this understanding and makes erroneous generalizations. 

LLaVA tends to do the opposite --- it misclassifies paintings important for the style they represent. For example, LLaVA mislabeled Cezanne's Post- Impressionism works as Impressionism. This can be considered an honest mistake, however, Cezanne played an important role for Post-Impressionism. On the other hand, LLaVA labeled Realism paintings by Courbet as Romanticism, which is odd given that Courbet rebelled against Romanticism. Interestingly, LLaVA states a `focus on emotions and individualism', which is true for both Realism and Romanticism, but in a different manner. In general, these models fail to consider all relevant stylistic and contextual elements to reason about the correct style for works that are harder to classify.

\subsubsection{Adding Information in the Prompt}

For LLaVA, we find that adding author and creation year information improves the style classification performance over the baseline with no additional information in JenAesthetics 
(Table~\ref{tab:LLaVA_prompts}).
This suggests that LLaVA has learned correlations between historical periods and art styles. In addition, it shows that LLaVA recognizes the established painters in the evaluated datasets. Both author and year information improve the accuracy, while providing both 
information jointly results in the highest accuracy, with an increase of up to 18\% on JenAesthetics compared to the baseline.
For WikiArt, however, additional information does not improve art style classification. This might be due to WikiArt containing paintings of varying popularity, such as those from Russian artists Nicholas Roerich, Pyotr Konchalovsky, Boris Kustodiev, 
or Zinaida Serebriakova, who might be less known by an international audience and infrequently discussed online. 

\begin{table}[b]
\caption{Zero-shot art style prediction by LLaVA for different input prompts on WikiArt (WA) Validation (V) and Validation Subset (VS) (for the predictions involving the year in the prompt) and JenAesthetics.}
\centering
\resizebox{\textwidth}{!}{%
\begin{tabular}{ l | c | c }
\toprule
\multicolumn{1}{l}{\textbf{LLaVA Prompt}} & \textbf{WikiArt V/VS} \newline \footnotesize{(Accuracy(\%))} & \textbf{JenAesthetics} \newline \footnotesize{(Accuracy(\%))} \\
\midrule
\textit{To which art (style)/(period) does this painting belong?} & 28.32 & 37.02 \\
\textit{What is the art (style)/(period) of this painting by [author]?} & 26.33 & 48.03 \\
\textit{What is the art (style)/(period) of this painting made in [year]?} & 29.55 & 50.89 \\
\textit{What is the art (style)/(period) of this painting made by [author] in [year]?} & 23.13 & 55.01 \\
\bottomrule
\end{tabular}%
}
\label{tab:LLaVA_prompts}
\end{table}

\subsection{Zero-Shot Author Prediction}

We present zero-shot author prediction results in Table~\ref{tab:author_prediction}.
All VLMs, including the low accuracy of 0.43\% on WikiArt by LLaVA, perform significantly above random chance. This equals to 0.092\% and 0.25\% on WikiArt validation set and JenAesthetics, which contain artworks from 1077 and 405 painters. For JenAesthetics, CLIP's most predicted author is Charles Francois Daubigny, who mainly painted landscapes, using a not very pronounced style spanning Romanticism, Impressionism, and Realism. The high frequency of Daubigny predictions might be explained by the wide range of styles that exist in close proximity to each other, the popularity of the theme of landscape, and the absence of key features that define a personal style. Notably, Pablo Picasso is not among CLIP's most predicted authors, despite being one of the most prevalent artists in WikiArt. 

\begin{table}[!t]
\caption{Zero-shot author prediction accuracy by CLIP, LLaVA and GPT-4o on WikiArt Validation and JenAesthetics. Note: We did not prompt GPT-4o to predict authorship; instead, we retrieved this information from the style predictions.}

\centering
\fontsize{8}{8}\selectfont
\resizebox{\textwidth}{!}{%
\begin{tabular}{l|l|c|c}
\toprule
\textbf{Model} & \textbf{Prompt} & \begin{tabular}[c]{@{}c@{}}\textbf{WikiArt Valid.}\\ \textbf{Accuracy (\%)}\end{tabular} & \begin{tabular}[c]{@{}c@{}}\textbf{JenAesthetics}\\ \textbf{Accuracy (\%)}\end{tabular} \\
\midrule
CLIP & \textit{this painting was made by [author]} & 26.39 & 36.23 \\
LLaVA & \textit{Who is the author of this painting?} & 4.35 & 4.95 \\
\midrule
GPT-4o & \textit{To which art (style)/(period) does this painting belong?} & 6.74 & 12.57 \\
\bottomrule
\end{tabular}%
}
\label{tab:author_prediction}
\end{table}

LLaVA outputs `unknown' in 41\% of times in JenAesthetics and 47.9\% of times in WikiArt. It also predicts only a limited set of author names, among which Vincent Van Gogh, Claude Monet, John Singer Sargent, Jackson Pollock, Michelangelo, John Constable, Pablo Picasso, Erro, John William Waterhouse.


\subsection{Zero-Shot Time Period Prediction}

Table~\ref{tab:year_prediction} shows the prediction results for decade and century. 
All VLMs perform better than a random classifier, except for LLaVA on WikiArt VS.
When asked, \textit{``When was this painting made?''}, LLaVA often responses such as ``\textit{The painting was made in the 16th century.}''. We convert these predictions to a rounded year, such as 1500. Since LLaVA often predicts only the century, the decade accuracies are low. On WikiArt VS, LLaVA often predicts `19th Century', which we convert to 1800. However, this prediction is incorrect, since all paintings with a year label in WikiArt VS are dated after 1900. Due to frequent 19th Century predictions, LLaVA performs worse than random on century prediction in WikiArt.

\subsection{A Closer Look at GPT-4o}

Due to its cost, we tested GPT-4o more selectively and only asked for style predictions. Nevertheless, when prompted to predict the art style, GPT-4o provides detailed descriptions, often mentioning the historical period and author. The time period is typically mentioned when describing an art style, most commonly as a century interval, in 90.84\% of WikiArt and 89.84\% of JenAesthetics predictions. We evaluate the accuracy of these descriptions by comparing statistics on the frequently mentioned authors with the ground truth.

\begin{table}[!t]
\caption{Zero-shot time period prediction accuracy by CLIP and LLaVA on WikiArt Validation Subset (VS) and JenAesthetics. We include the average performance of random predictions drawn from a uniform distribution.}
\centering
\fontsize{7}{8}\selectfont
\resizebox{\textwidth}{!}{%
\begin{tabular}{l|l|cc|cc}
\toprule
\textbf{Model} & \textbf{Prompt} & 
\multicolumn{2}{c|}{\textbf{WikiArt (VS)}} & 
\multicolumn{2}{c}{\textbf{JenAesthetics}} \\
\cmidrule(r){3-4} \cmidrule(l){5-6}
& & \textbf{Century (\%)} & \textbf{Decade (\%)} & \textbf{Century (\%)} & \textbf{Decade (\%)} \\
\midrule
CLIP & \textit{this painting was made in [year]} & 89.25 & 25.34 & 51.02 & 10.98 \\
LLaVA & \textit{When was this painting made?} & 36.65 & 10.85 & 63.43 & 2.73 \\
\midrule
\multicolumn{2}{l|}{Random classifier} & 87.49 $\pm$ 0.34 & 8.9 $\pm$ 0.3 & 18.66 $\pm$ 0.97 & 1.98 $\pm$ 0.33 \\
\bottomrule
\end{tabular}%
}
\label{tab:year_prediction}
\end{table}

In WikiArt, an author is mentioned in 12.06\% of the style descriptions. Appendix C lists the most frequently predicted authors. Multiple authors are mentioned concurrently to describe a style, e.g., Vincent van Gogh, Paul Cézanne, and Paul Gauguin are cited as key figures of Post-Impressionism. Pablo Picasso and Georges Braque are often mentioned as pioneers of Cubism. 
In JenAesthetics, authors appear in 24.12\% of the paintings. This higher rate can be due to paintings in JenAesthetics being mostly renown, Western artworks, possibly reflecting training bias in GPT-4o. 
The top five mentioned authors are Henri Rousseau, Claude Monet, Paul Cézanne, Rembrandt, Vincent Van Gogh -- all pioneers in their periods. 
These are correct for 6.74\% (WikiArt) and 12.57\% (JenAesthetics), with the latter notably outperforming LLaVA (Table 3).

\section{Conclusions}
We acknowledge several benefits of VLMs over earlier artwork classifiers, which are typically trained solely for art style classification and cannot predict other art historical attributes without fine-tuning. In contrast, we show that a single VLM can perform three tasks -- predicting art style, author and time period -- above random chance, without specific training. 

We focus on zero-shot performance to provide a foundational understanding of VLMs. This approach reflects typical real-world use by non-experts and tests whether VLMs' generalization across image domains extends to artworks. Although less accurate than models fine-tuned for style classification, VLMs provide insights into their reasoning. Artistic styles often share overlapping characteristics. Rather than strictly adhering to predefined labels, VLMs occasionally make misclassifications that align with stylistic similarities, indicating an emerging grasp of nuanced artistic traits. Notably, LLaVA and GPT-4o generate detailed explanations of features that define a painting's style. These explanations enhance understanding and clarify classifications, offering greater transparency compared to traditional deep learning models.

Nevertheless, we find that VLMs, even the powerful GPT-4o, make mistakes that do not align with expert art historical standards. They may misclassify even canonical artworks by renowned artists or provide flawed reasoning. We conclude that VLMs, despite their great potential, have not yet mastered art history. Their limitations pose risks for non-experts, who may rely on these models, as they could assign incorrect styles, authors, or time periods to lesser-known or unknown pieces. This highlights the importance of expert oversight and the involvement of art historians in verifying and refining AI-generated information.

%
%
\bibliographystyle{splncs04}
\bibliography{main}

\section*{Appendices}


\appendix

\section{Example of artworks for each style in WikiArt}\label{Appendix A}

\begin{figure}[htbp]
    \centering
    \includegraphics[width=1.0\textwidth]{figures_new/WikiArt.pdf}
    \caption{
    Example of artworks for each style in WikiArt, sorted chronologically. From top left, 
    \textit{Baptism of Christ} by P. della Francesca,
    \textit{The Arnolfini Portrait} by J. van Eyck,
    \textit{The Garden of Earthly Delights} by H. Bosch,
    \textit{The Holy Trinity} by El Greco,
    \textit{Las Meninas} by D. Velázquez,
    \textit{The swing} by J. Fragonard,
    \textit{Saturn Devouring His Son} by F. Goya,
    \textit{The Waterfall Where Yoshitsune Washed His Horse at Yoshino in Yamato Province} by K. Hokusai,
    \textit{A widow} by J. Tissot,
    \textit{Rouen Cathedral, the Portal in the Sun} by C. Monet,
    \textit{Capo di Noli} by P. Signac,
    \textit{Clownesse Cha-U-Kao} by H. de Toulouse-Lautrec,
    \textit{The tragedy} by P. Picasso,
    \textit{The kiss} by G. Klimt,
    \textit{Charing Cross Bridge} by A. Derain,
    \textit{Portrait of a woman} by G. Braque,
    \textit{Les Demoiselles d'Avignon} by P. Picasso,
    \textit{Grapes} by J. Gris,
    \textit{Figures Lying on the Sand} by S. Dalí,
    \textit{The sleeping gypsy} by H. Rousseau,
    \textit{Woman I} by W. de Kooning, 
    \textit{New York} by F. Kline, 
    \textit{Color Maze} by F. Stella
    \textit{Cape Cod Morning} by E. Hopper,
    \textit{Crying girl} by R. Lichtenstein 
    \textit{Immaginando Tutto} by A. Boetti 
    and 
    \textit{Anne in a Striped Dress} by F. Porter.
    }
    \label{fig:art_styles}
\end{figure}

\newpage  
\section{Related Work}\label{Appendix B}

For comparison with the VLMs performance, we provide an overview of the style classification accuracies achieved on WikiArt by previous work in Table~\ref{tab:WikiArt_baselines}.
    
\begin{table}[h]
\caption{Art style prediction performance of previous works on WikiArt.}
\centering
\fontsize{10}{10}\selectfont
\begin{tabular}{l|ll}
\toprule
 Method & Acc. (\%) & \# Classes \\
\midrule
Karayev et al. \cite{karayev2013recognizing} & 44.10 &  25 \\
Bar et al. \cite{bar2015classification} & 43.02 & 27 \\
Saleh and Elgammal \cite{saleh2015large} & 46.00 & 27 \\
Tan et al. \cite{tan2016ceci} & 54.50 & 27 \\
Lecoutre et al. \cite{lecoutre2017recognizing} & 62.80 & 25 \\
Florea and Gieseke \cite{florea2018artistic} & 46.20 & 25 \\
Cetinic et al. \cite{cetinic2018fine} & 56.40 & 27 \\
Zhong et al. \cite{zhong2020fine} & 59.01 & 25 \\
Sandoval et al. \cite{sandoval2019two} & 66.71 & 22 \\
Menis-Mastromichalakis et al. \cite{Menis2020Deep} & 68.55 & 21 \\
Zhao et al. \cite{zhao2022big} & 71.24 & 27 \\
\bottomrule
\end{tabular}
\label{tab:WikiArt_baselines}
\end{table}

\section{Styles predicted by GPT-4o compared to the ground art-styles in JenAesthetics}\label{Appendix C}

\begin{figure}[h]
    \centering
    \includegraphics[width=1\textwidth]{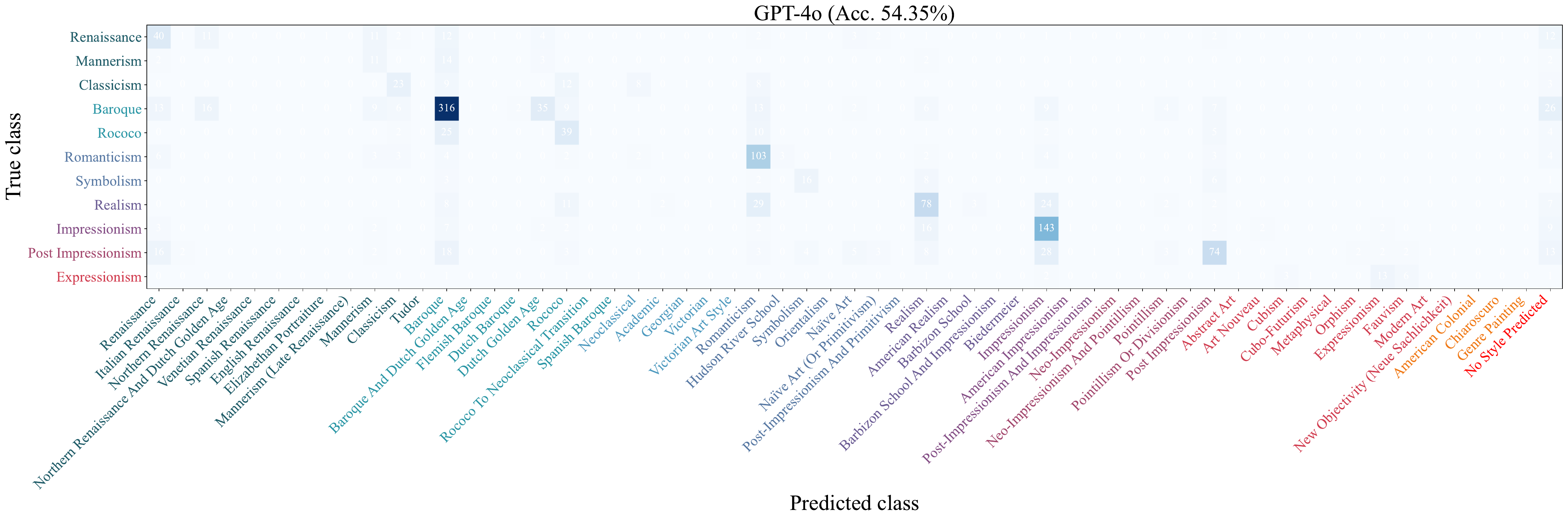}
    \caption{Styles predicted by GPT-4o compared to the ground art-styles in JenAesthetics.
    The color codes denote the overarching artistic periods:
    \textcolor{DarkerTealBlue}{Renaissance}, 
    \textcolor{CyanBlue}{Baroque and Rococo}, 
    \textcolor{SkyBlue}{Neoclassicism}, 
    \textcolor{SlateBlue}{Romanticism, Symbolism and Naïve Art/Primitivism}, 
    \textcolor{Indigo}{Realism}, 
    \textcolor{DarkPurple}{Impressionism}, 
    \textcolor{MagentaPurple}{Post-Impressionism}, 
    \textcolor{CrimsonRed}{Modern Art and 20th Century Movements} and \textcolor{DeepOrange}{Others} (territorial labels/techniques/genre). Styles in the same period share visual and technical similarities. GPT-4o provides fine-grained style classifications compared to the dataset's ground-truth labels.}
    \label{fig:JenAesthetics_GPT-4o}
\end{figure}

\section{Zero-shot Art Style classification}\label{Appendix D}

\subsection{Dataset Statistics}
\begin{figure}[htbp]
    \centering
    \includegraphics[width=1\textwidth]{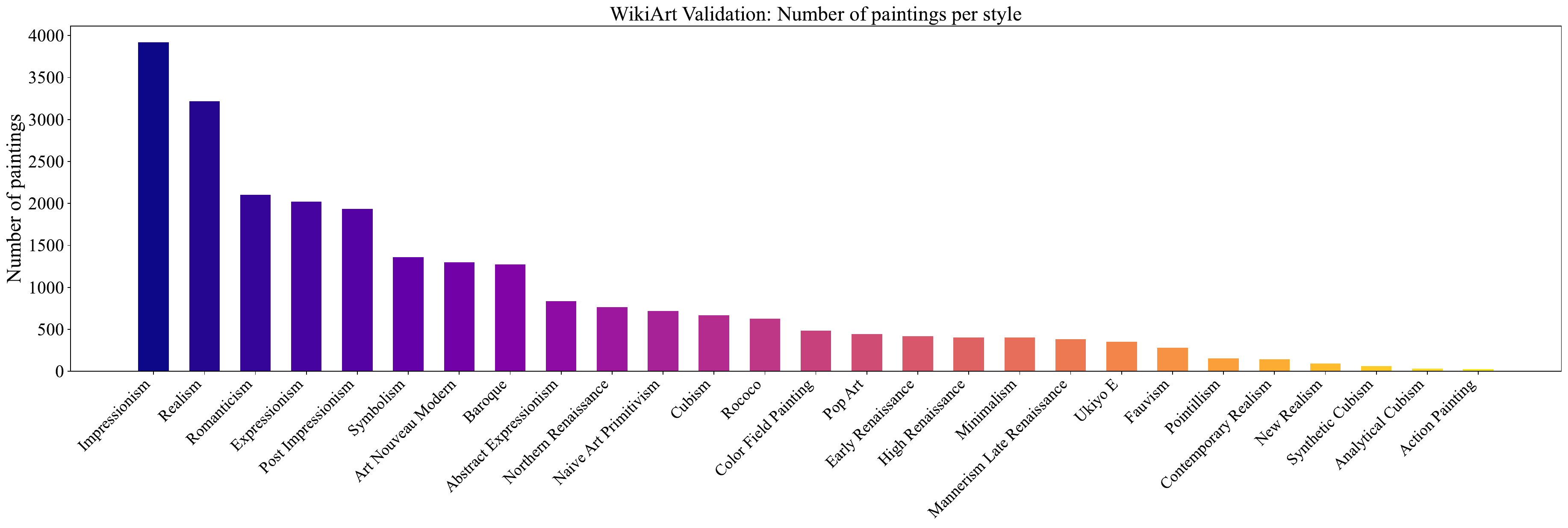}
    \includegraphics[width=1\textwidth]{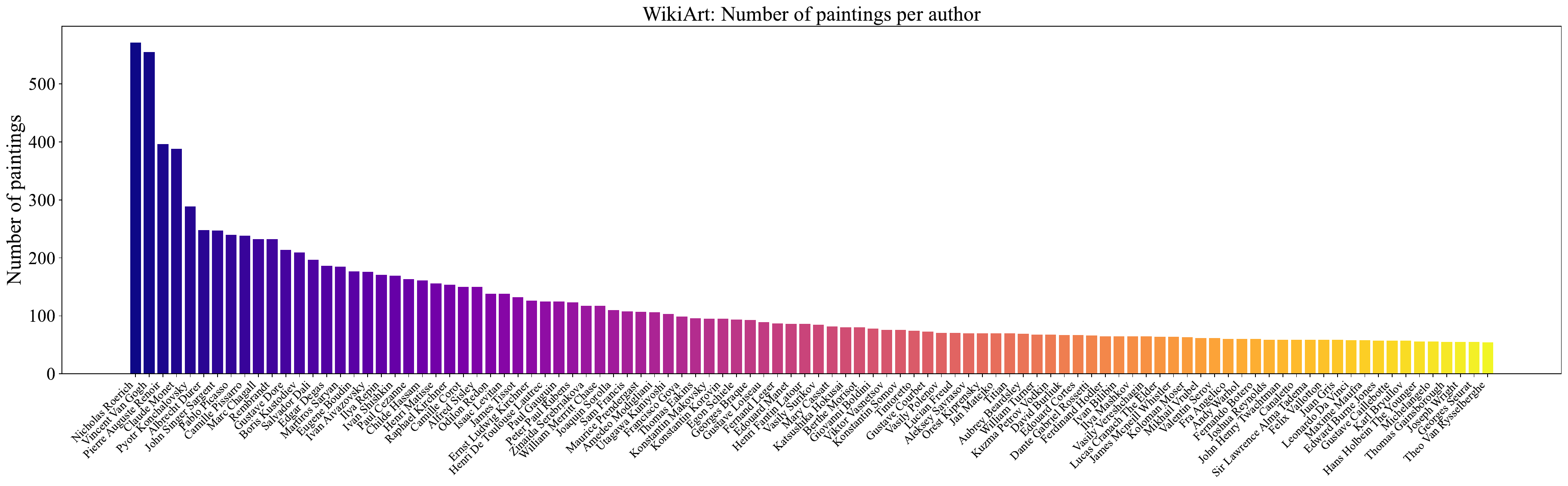}
    \includegraphics[width=1\textwidth]{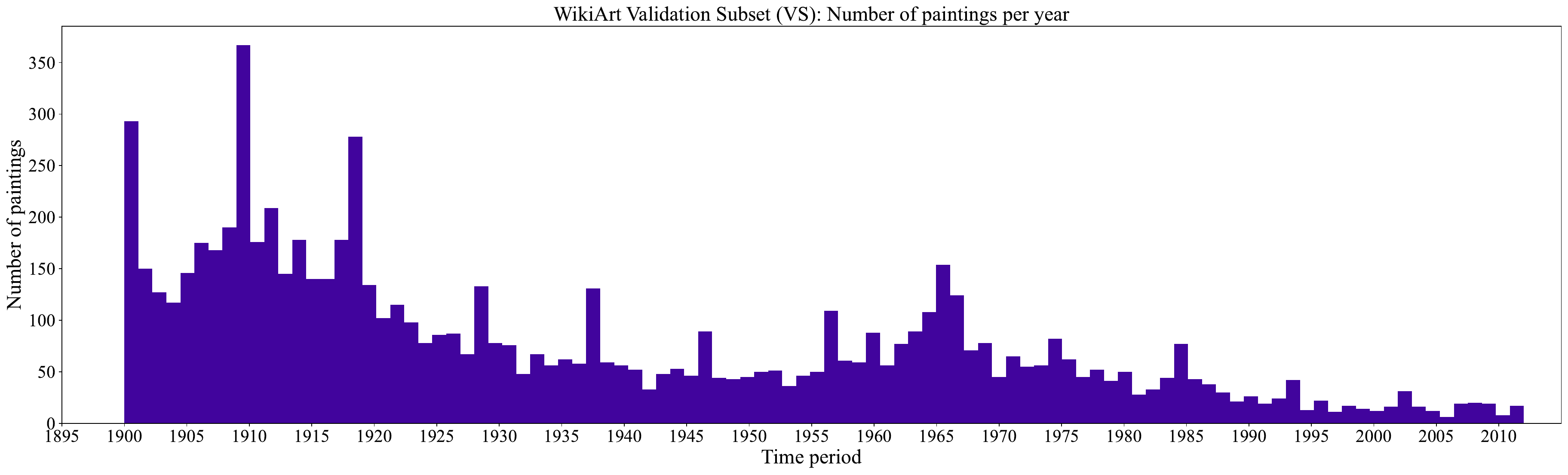}
    \caption{Number of paintings per art style (top), artist (middle) and time period (bottom) in the WikiArt Validation set and WikiArt Validation Subset (VS). For visualization purposes, we only show the 100 most frequent artists.}
    \label{fig:WikiArt_stats}
\end{figure}

\begin{figure}[h]
    \centering
    \includegraphics[width=1\textwidth]{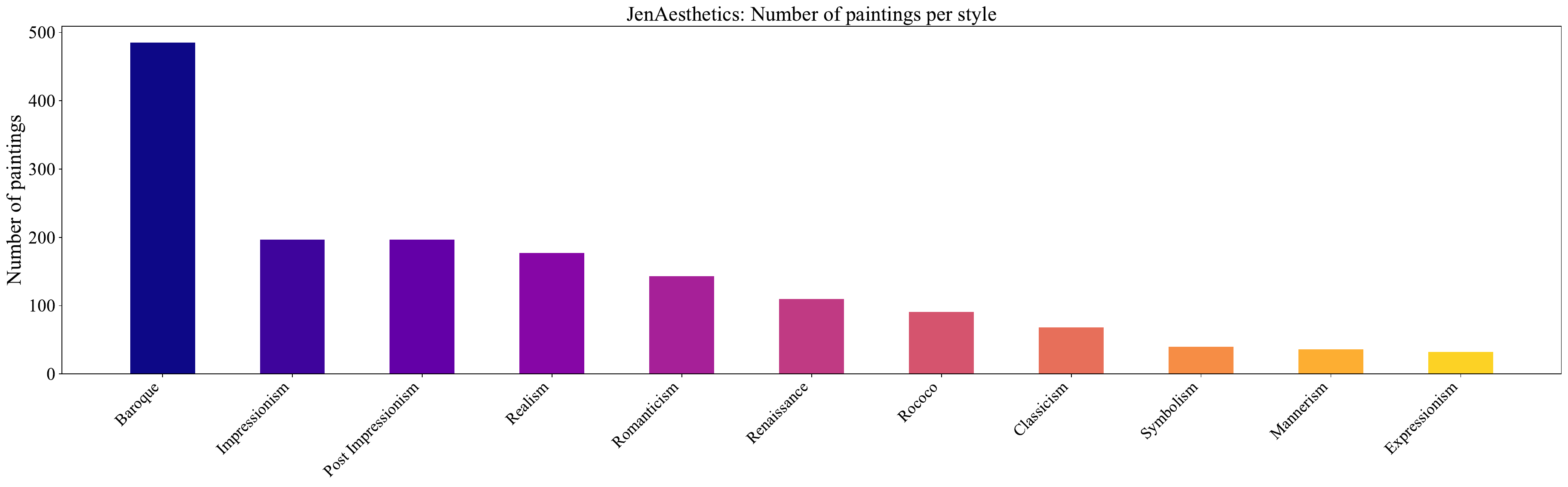}
    \includegraphics[width=1\textwidth]{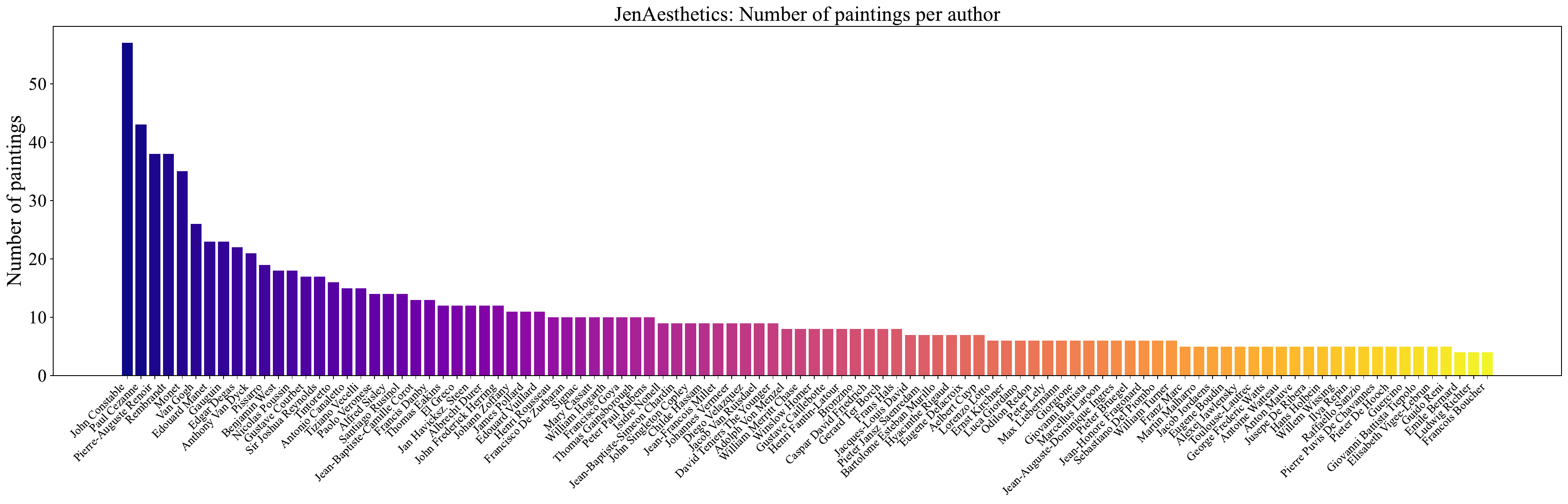}
    \includegraphics[width=1\textwidth]{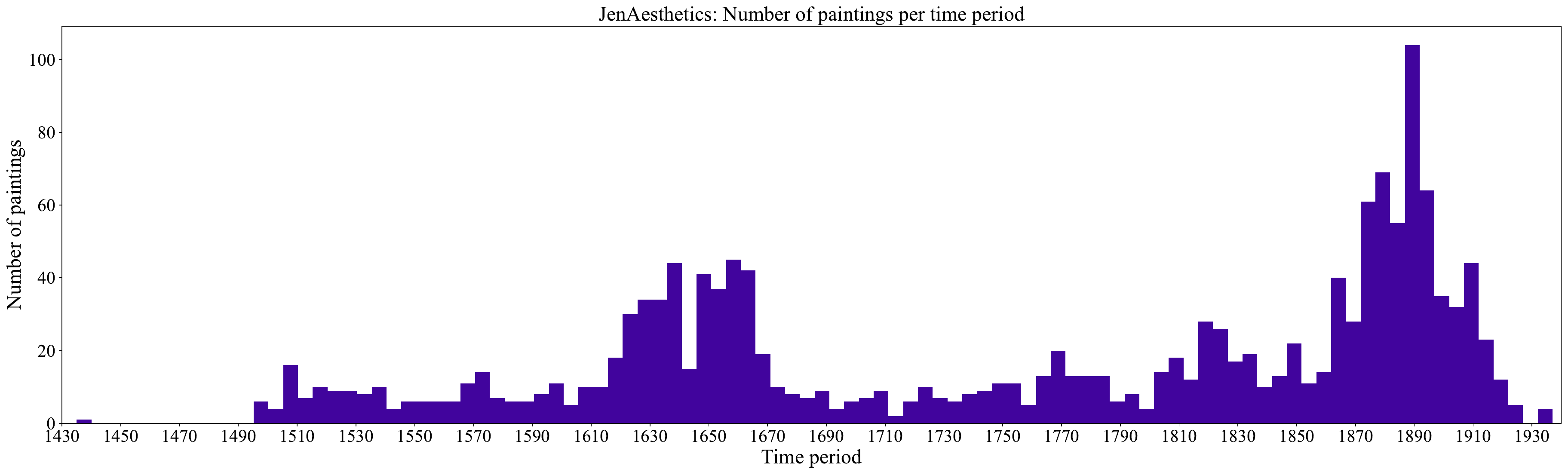}
    \caption{Number of paintings per art style (top), artist (middle) and time period (bottom) in the JenAesthetics dataset. For visualization purposes, we only show the 100 most frequent artists.}
    \label{fig:JenAesthetics_stats}
\end{figure}

We report the number of paintings by art style, artist, and time period for WikiArt and JenAesthetics in Figures~
\ref{fig:WikiArt_stats}, \ref{fig:JenAesthetics_stats}.

\clearpage  
\subsection{Precision and Recall}

\begin{figure}[h]
    \centering
    \includegraphics[width=1\textwidth]{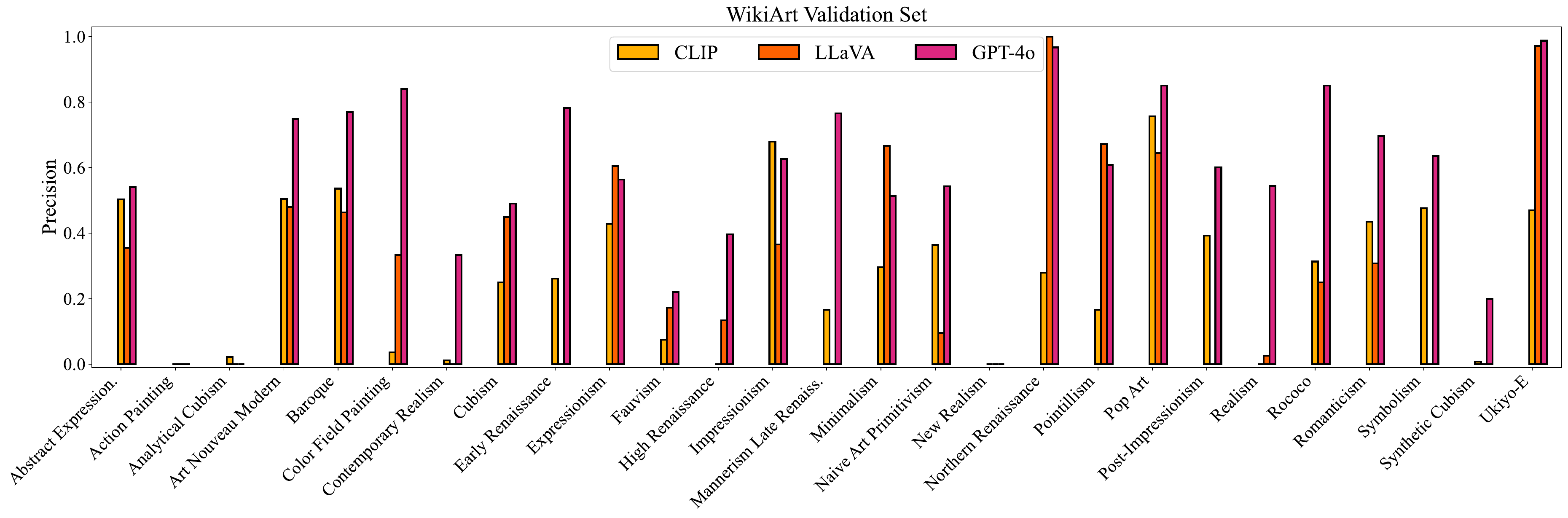}
    \includegraphics[width=1\textwidth]{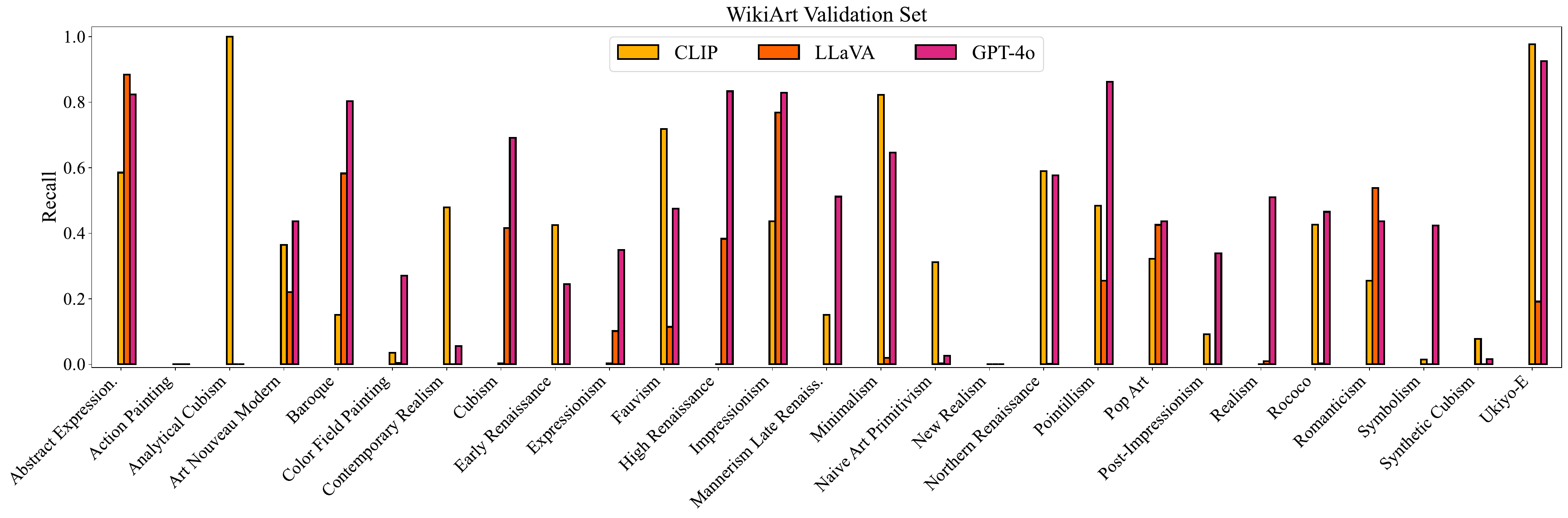}
    \includegraphics[width=1\textwidth]{figures_new/WikiArt_F1_score_bars.pdf}
    \caption{Precision, Recall and F1 score for the different art styles in the WikiArt Validation set.}
    \label{fig:PRF_WikiArt}
\end{figure}

\begin{figure}[ht!]
    \centering
    \includegraphics[width=1\textwidth]{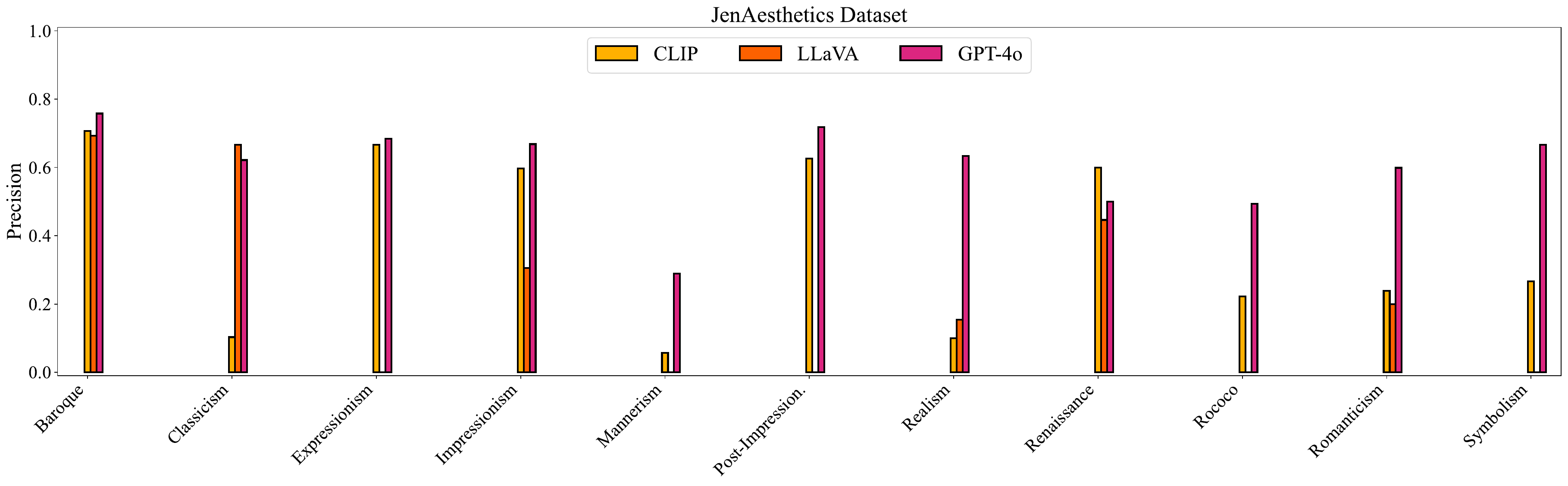}
    \includegraphics[width=1\textwidth]{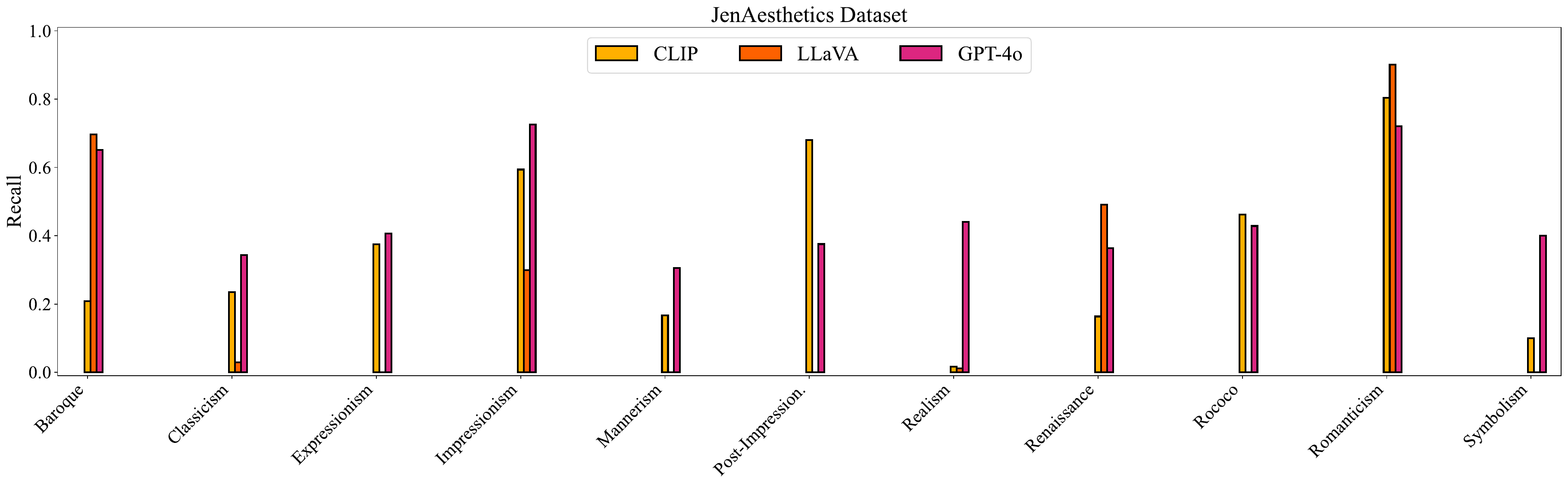}
    \includegraphics[width=1\textwidth]{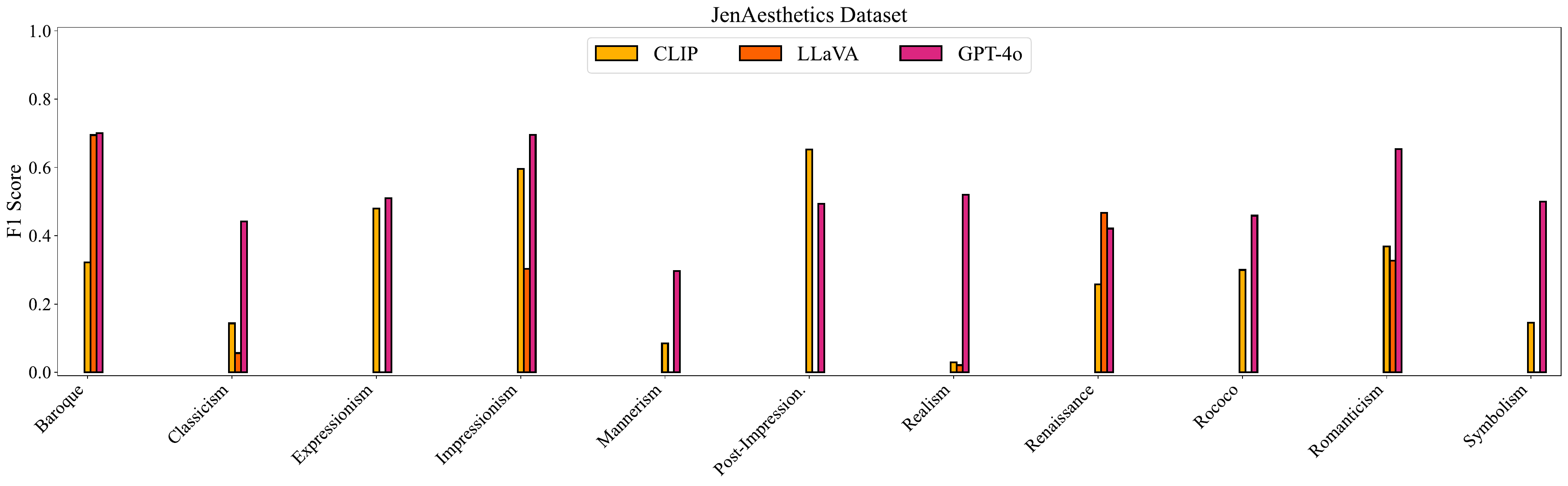}
    \caption{Precision, Recall and F1 score for the different art styles in the JenAesthetics dataset.}
    \label{fig:PRF_JenAesthetics}
\end{figure}

The precision, recall and F1 Score for the classification of the different styles are shown in Figures~\ref{fig:PRF_WikiArt}, and \ref{fig:PRF_JenAesthetics}.

\clearpage 
\subsection{Misclassifications}
\begin{figure*}[ht!]
    \centering   
    \includegraphics[width=0.45\textwidth]{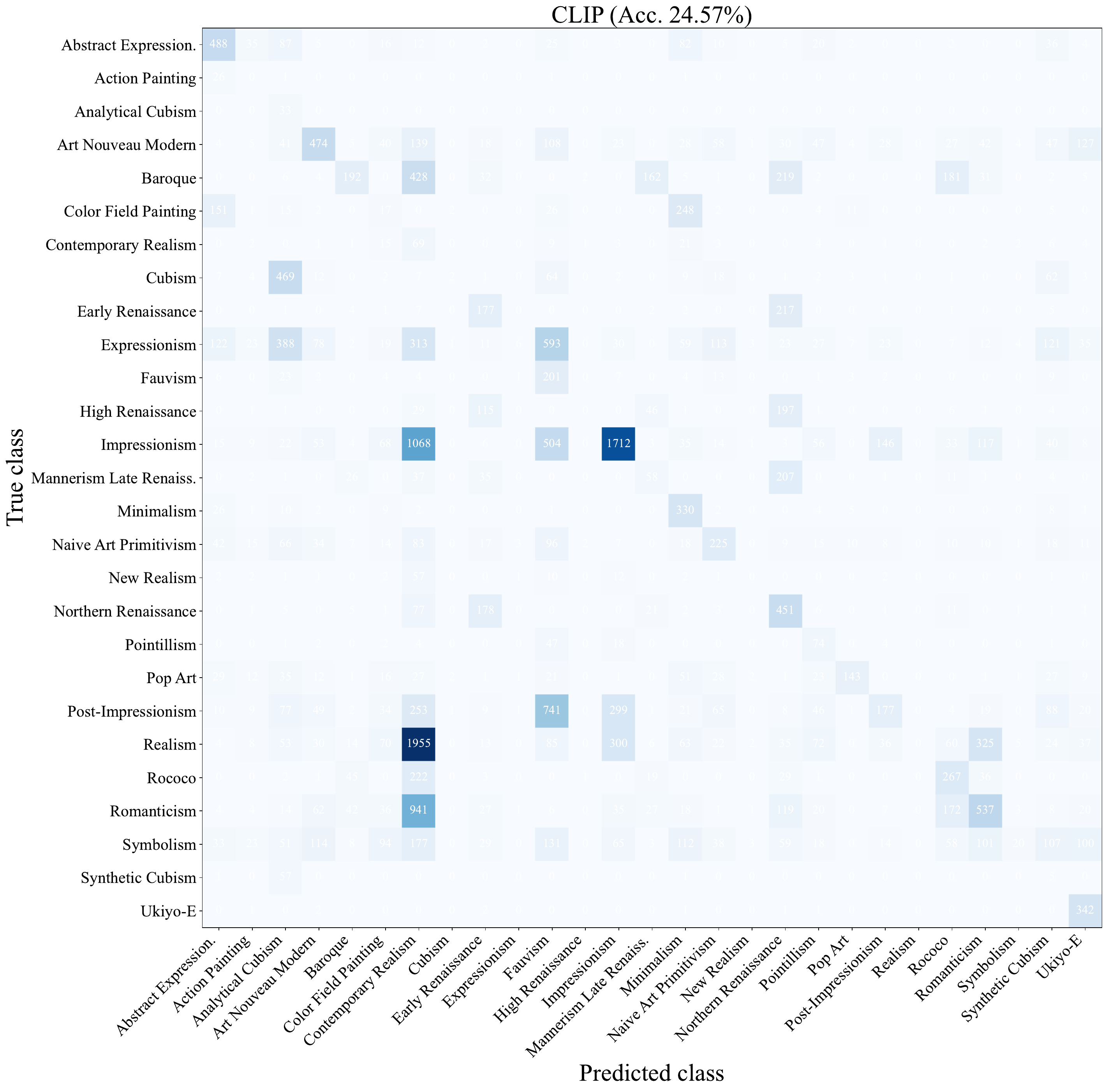}
    \includegraphics[width=0.95\textwidth]{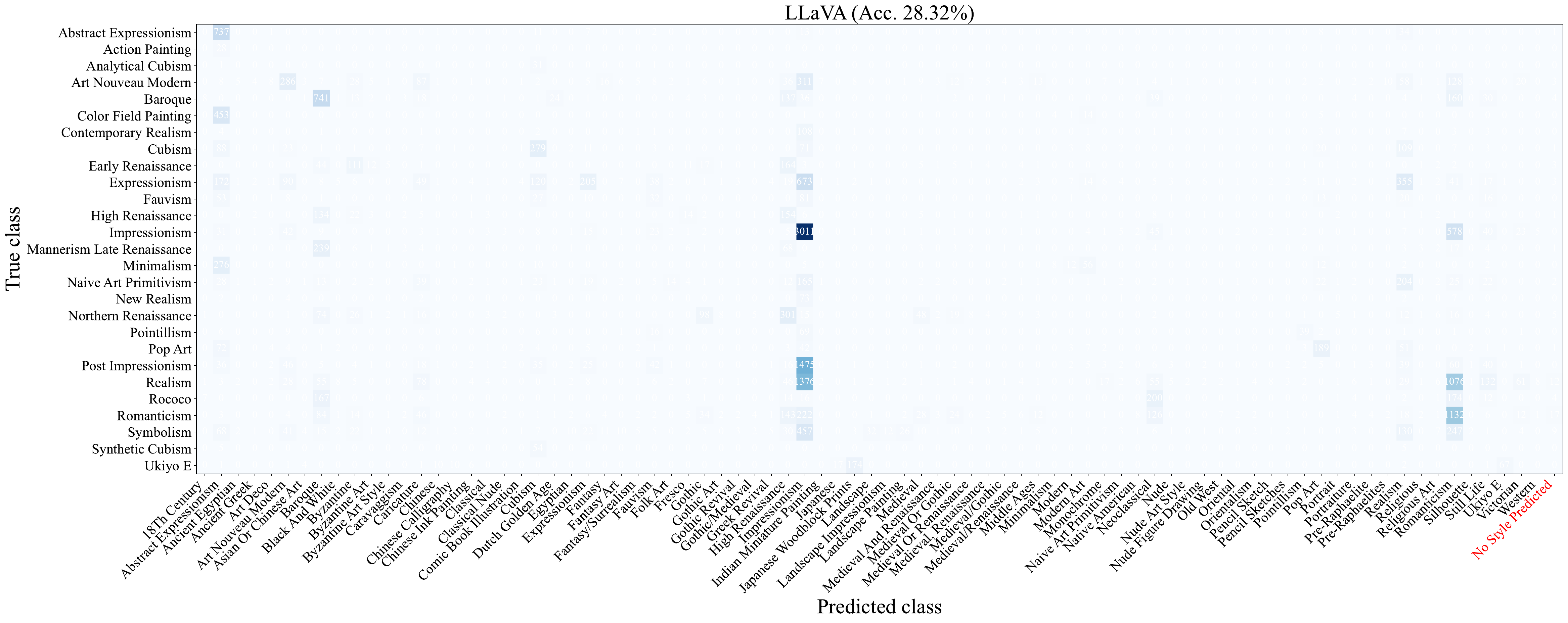}   
    \includegraphics[width=0.95\textwidth]{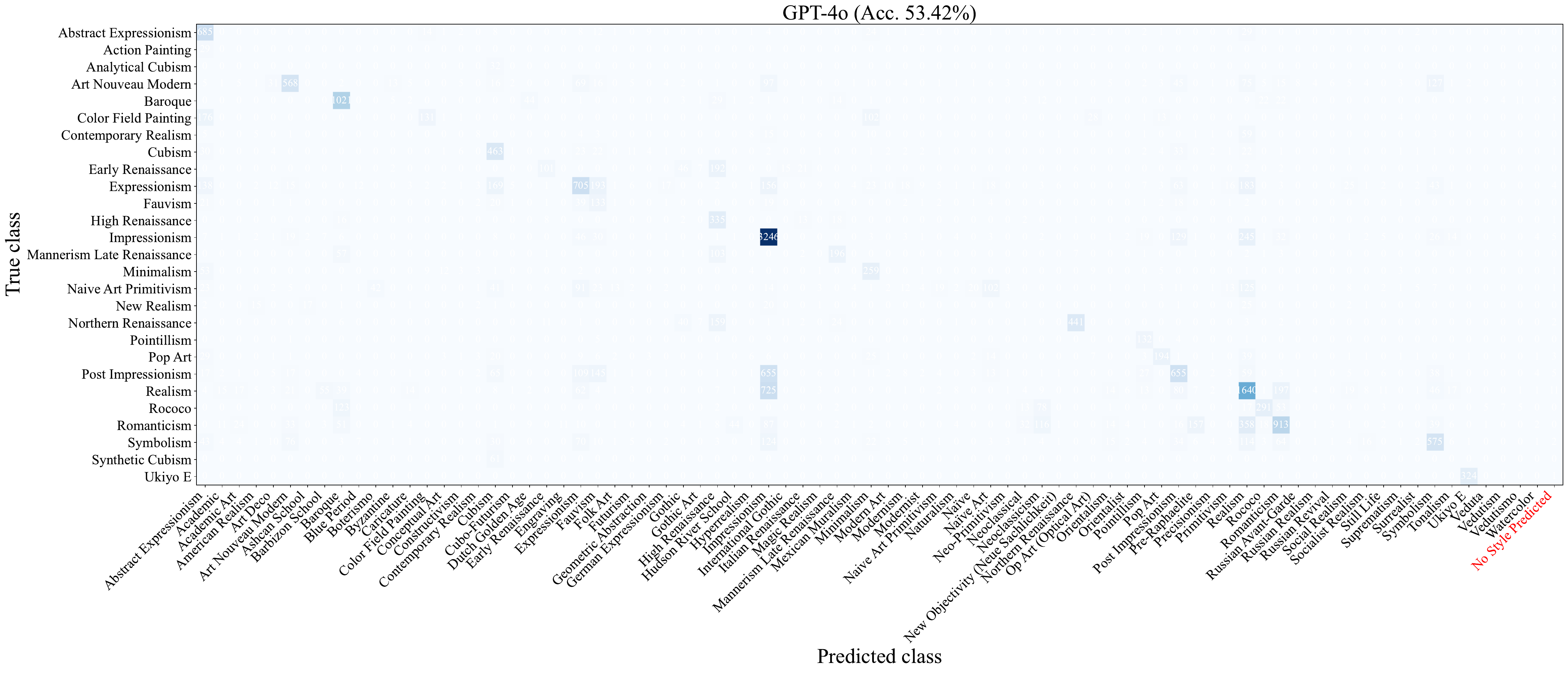}
    
    \caption{Predicted vs. ground-truth art styles for CLIP, LLaVA and GPT-4o for zero-shot style prediction on the WikiArt Validation set. For LLaVA and GPT-4o, only labels predicted at least 10 times are shown. }
    \label{fig:WikiArt_confusion}
\end{figure*}

\begin{figure}[ht!]
    \centering   
    \includegraphics[width=0.45\textwidth]{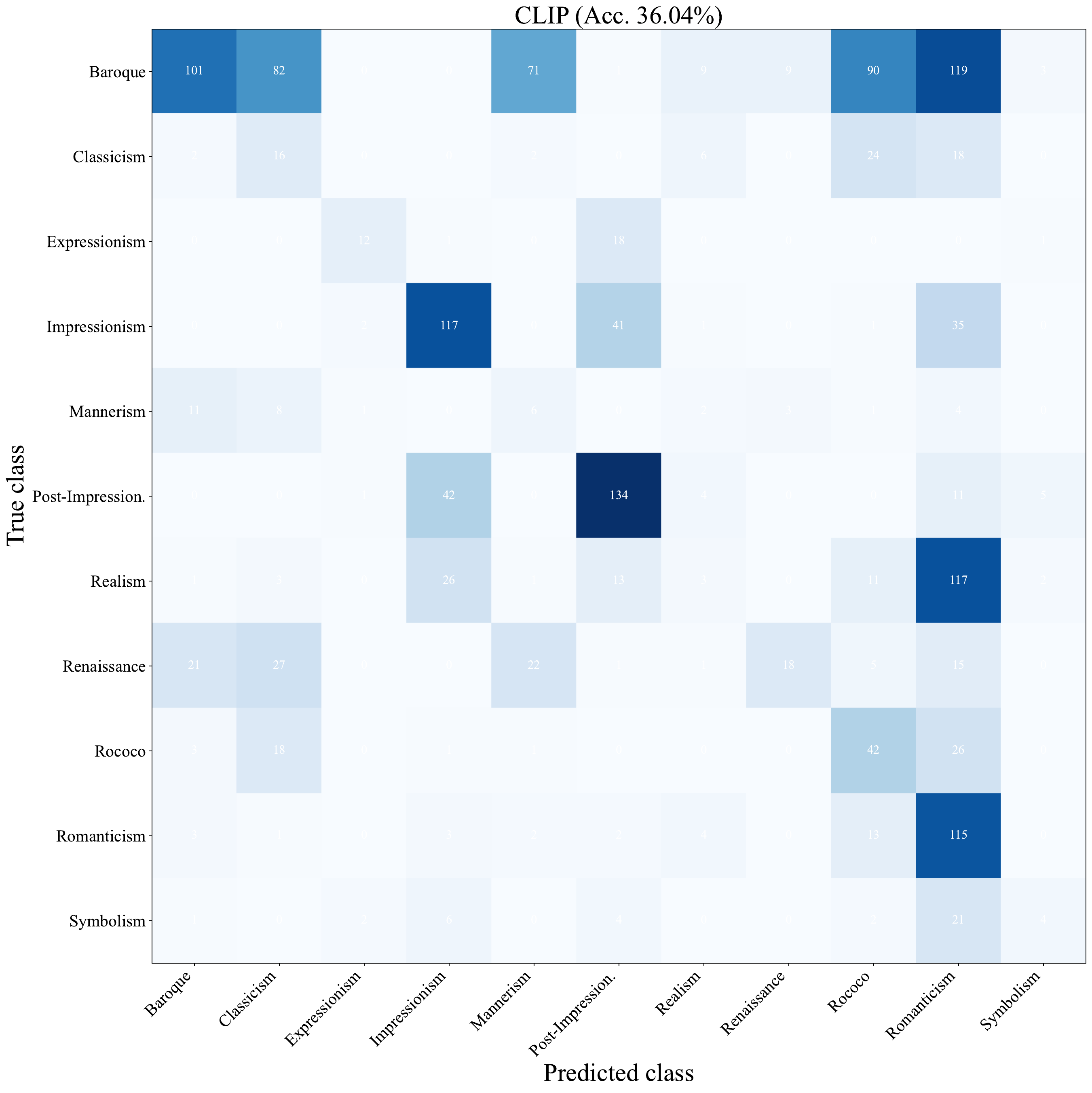}
    \includegraphics[width=0.95\textwidth]{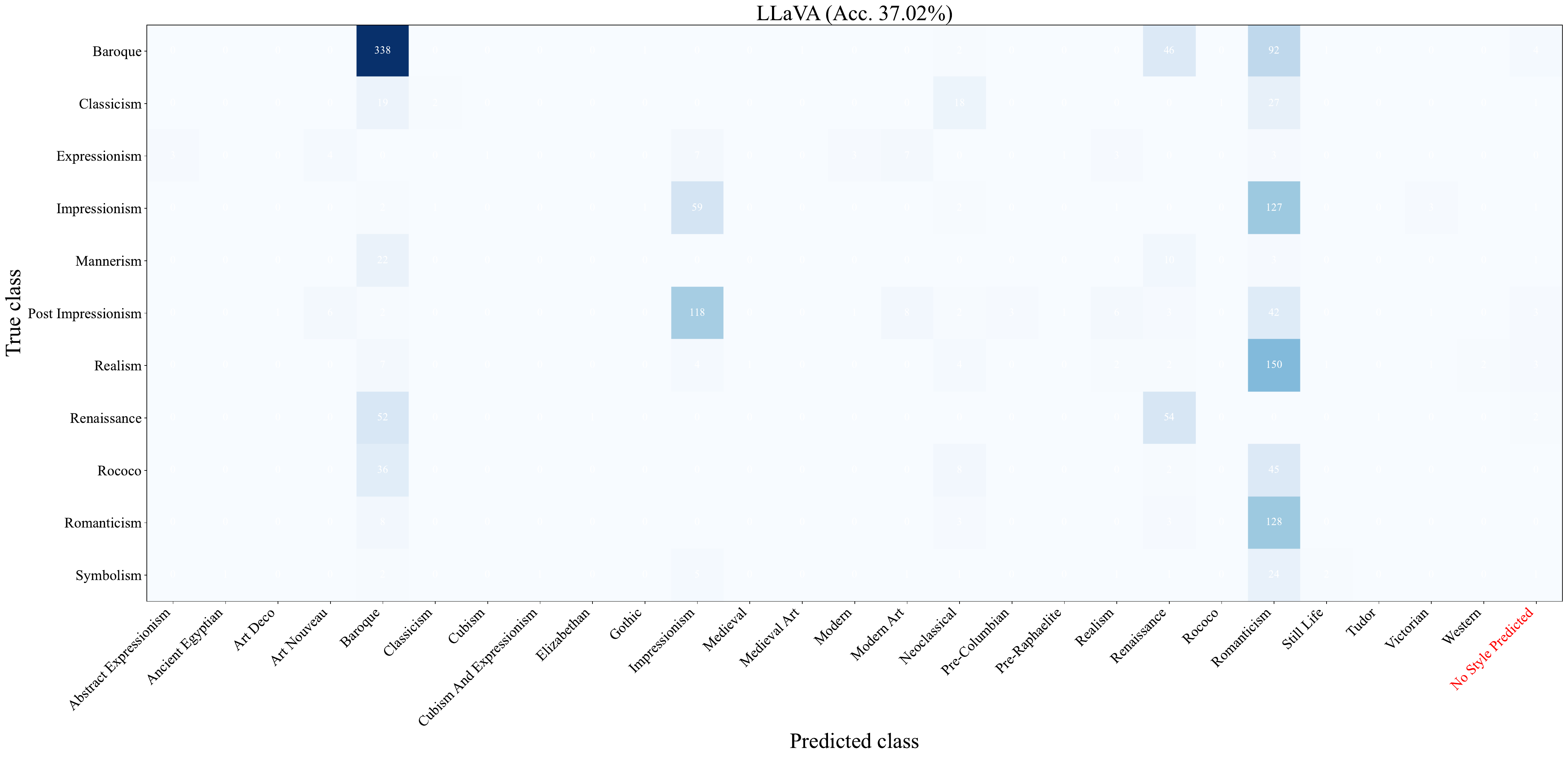}   
    \includegraphics[width=0.95\textwidth]{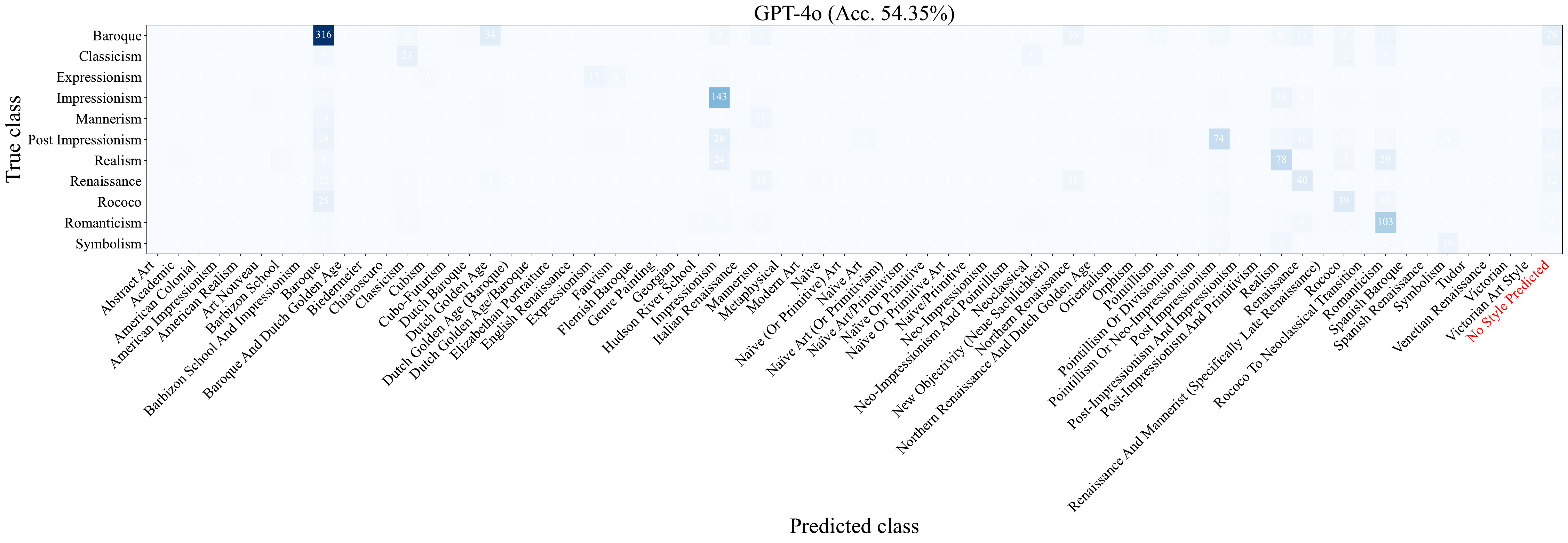}
    
    \caption{Predicted vs. ground-truth art styles for CLIP, LLaVA and GPT-4o for zero-shot style prediction the JenAesthetics dataset.}
    \label{fig:JenAesthetics_confusion}
\end{figure}

To better understand the misclassifications made by the evaluated VLMs, we plot the confusion matrices, available in Figures~\ref{fig:WikiArt_confusion} and \ref{fig:JenAesthetics_confusion}.

\clearpage 
\section{F1 Scores on WikiArt validation set}\label{Appendix F}

\begin{figure*}[ht!]
    \centering
    \includegraphics[width=1\textwidth]{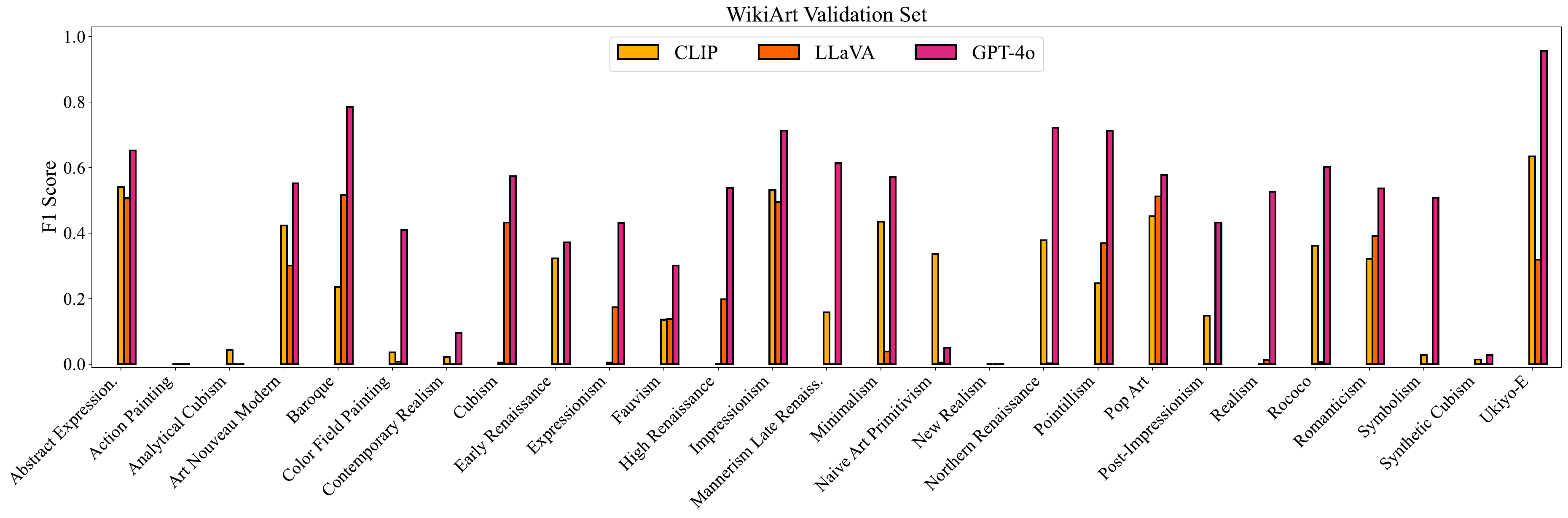}
    \caption{F1 Scores on WikiArt validation set for the classification of different art styles.
    }
    \label{fig:styles_precision_recall}
\end{figure*}

\section{A Closer Look at GPT-4o}\label{Appendix E}

\subsection{Art Style Descriptions}
\begin{figure}[htbp]
    \centering   
    \includegraphics[width=1.0\textwidth]{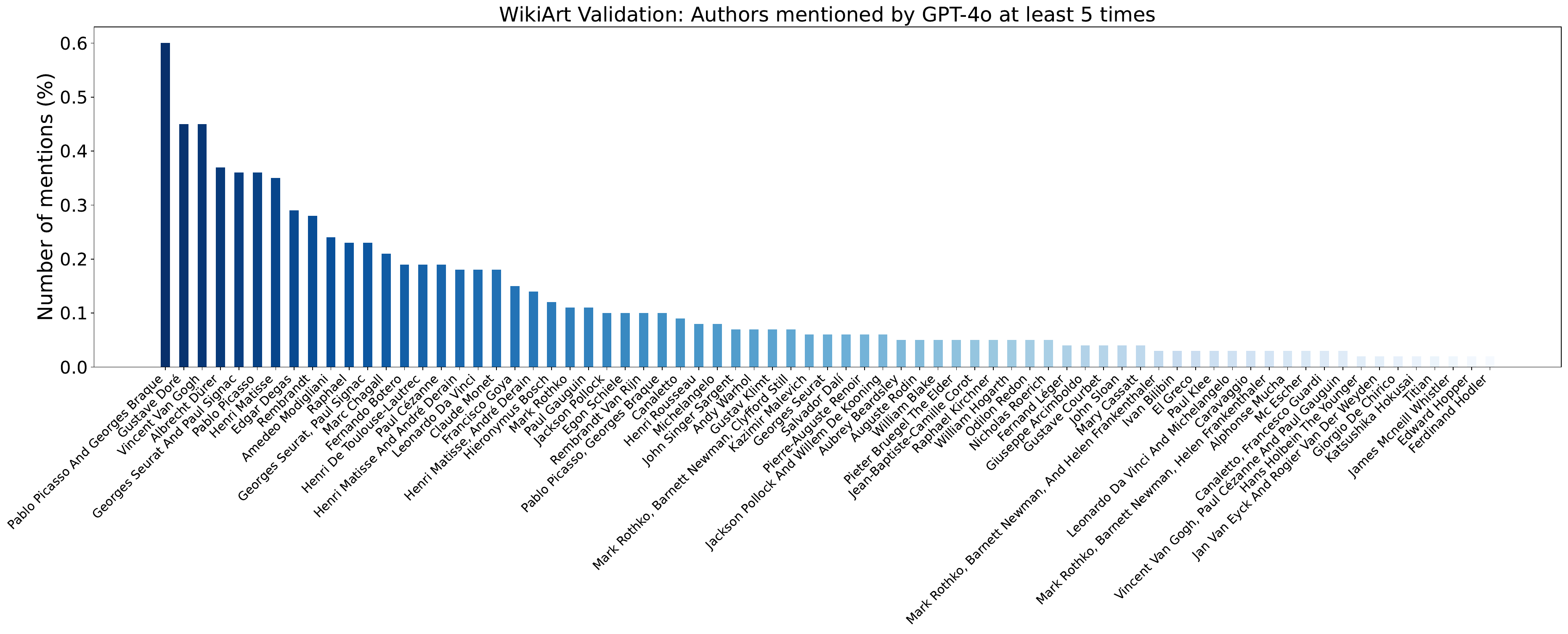}
    \includegraphics[width=1.0\textwidth]{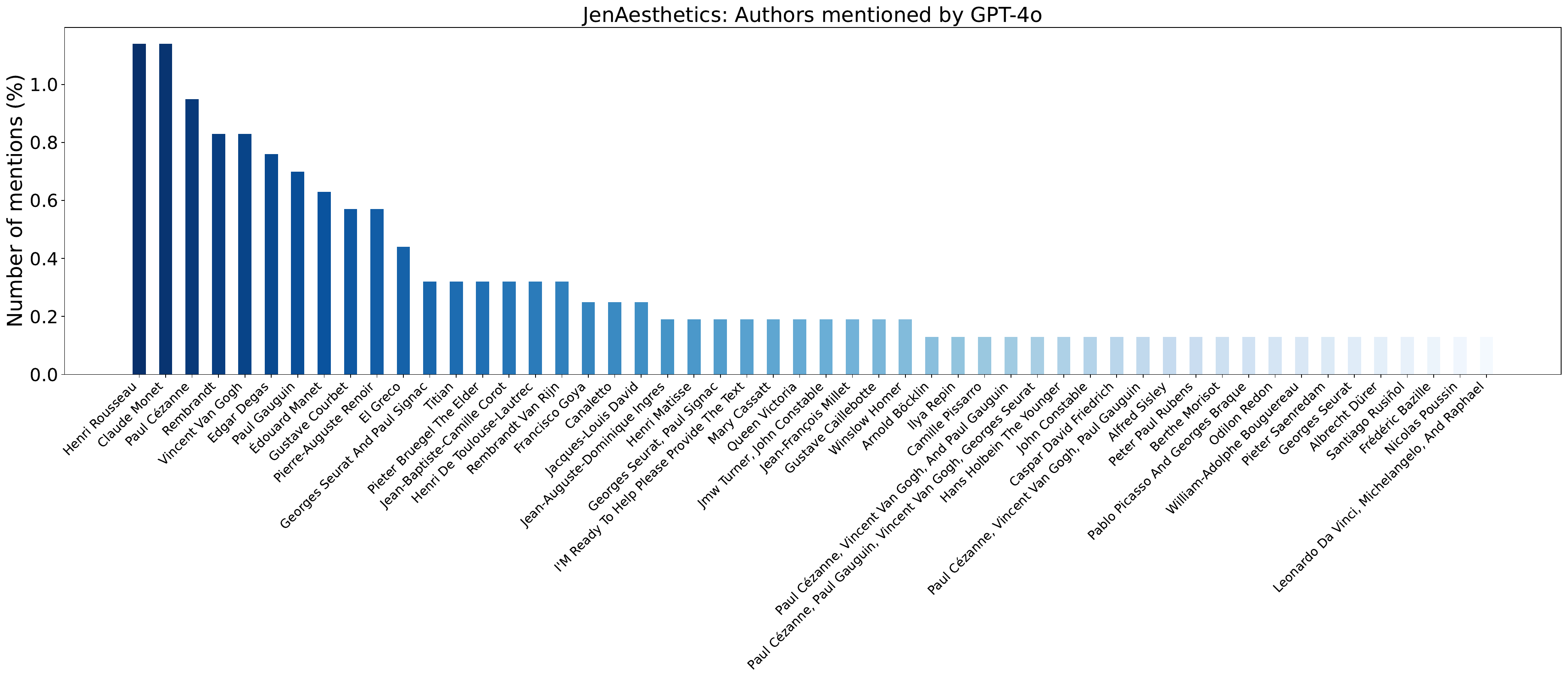}
    \caption{Painters mentioned by GPT-4o when prompted to predict the art style of a painting from the WikiArt Validation set and JenAesthetics datasets.}
    \label{fig:painter_mentions}
\end{figure}

When prompted to predict the art style of a painting, often GPT-4o provides rich descriptions of the painting characteristics and its artistic period. In some cases, these descriptions mention the name of the painter. The frequently mentioned painters are shown in Figure \ref{fig:painter_mentions}.

\clearpage 

\end{document}